# A Brief History of Digital Twin Technology

**Yunqi Zhang, PhD**

Department of Nuclear Medicine, Ruijin Hospital, Shanghai Jiao Tong University School of Medicine, 197 Ruijin Second Road, Shanghai 200025, China

College of Health Science and Technology, Shanghai Jiao Tong University School of Medicine, Shanghai, 200025, China

Email: zhangyunqi979@163.com

**Kuangyu Shi, PhD**

Department of Nuclear Medicine, Inselspital, Bern University Hospital, University of Bern, Bern 3010, Switzerland

Computer Aided Medical Procedures and Augmented Reality, Institute of Informatics, Technical University of Munich, Munich 80333, Germany

Email: kuangyu.shi@unibe.ch

**Corresponding Author**

**Biao Li, PhD, MD**

Department of Nuclear Medicine, Ruijin Hospital, Shanghai Jiao Tong University School of Medicine, 197 Ruijin Second Road, Shanghai 200025, China

College of Health Science and Technology, Shanghai Jiao Tong University School of Medicine, Shanghai, 200025, China

Email: lb10363@rjh.com.cn

**Synopsis**

Emerging from NASA's spacecraft simulations in the 1960s, digital twin (DT) technology has advanced through industrial adoption to spark a healthcare transformation. A DT is a dynamic, data-driven virtual counterpart of a physical system, continuously updated through real-time data streams and capable of bidirectional interaction. In medicine, DT integrates imaging, multi-omics, biosensors, and computational models to generate patient-specific simulations that support diagnosis, treatment planning, and drug development. Representative applications include cardiac DT for predicting arrhythmia treatment outcomes, oncology DT for tracking tumor progression and optimizing radiotherapy, and pharmacological DT for accelerating drug discovery through in silico trials. Despite rapid progress, major challenges, including interoperability, data privacy, and model fidelity, continue to limit widespread clinical integration. Emerging solutions such as explainable AI, federated learning, and harmonized regulatory frameworks offer promising pathways forward. Looking ahead, advances in multi-organ DT, genomics integration, and ethical governance will be essential to ensure that DT shifts healthcare from reactive treatment to predictive, preventive, and truly personalized medicine.

**Key points**

- Digital twin (DT) technology has expanded from aerospace applications to provide transformative tools in precision medicine, enabling patient-specific modeling in cardiology, oncology, and chronic disease management.
- The fusion of multiscale biological modeling, continuous real-time data from Internet of Things sensors, and artificial intelligence-driven analytics allows DT to simulate physiological processes from the molecular to the organ level, supporting personalized treatment optimization.
- Key challenges include data privacy, interoperability, and ensuring model accuracy and fidelity.
- Future DT development holds promise for proactive, predictive healthcare, shifting paradigms from reactive treatment to personalized prevention and early intervention.

**Introduction**

Digital twin (DT) technology represents a paradigm in cyber-physical systems, enabling real-time synchronization between physical entities and their virtual counterparts through dynamic, data-driven modeling (Fig. 1;Table 1)[1]. A DT can be defined as a virtual representation of a physical system, continuously updated with real-time data and capable of bidirectional interaction. This framework differs from traditional simulations by incorporating closed-loop feedback and lifecycle continuity, which allows DT to support decision-making across engineering and healthcare[2].

The concept was first introduced by Michael Grieves in 2002 as the "mirrored spaces model," later formalized by the United States National Aeronautics and Space Administration (NASA) in 2010 for spacecraft simulation[3,4]. NASA has defined DT as aircraft or system-oriented methods that make full use of the best physical models, sensors, and operating historical data, integrate multi-disciplinary and multi-scale probabilistic simulation processes, and map the state of a corresponding physical aircraft[5,6]. Subsequent advances in enabling technologies, including the Internet of Things (IoT) technology[7], Artificial Intelligence (AI)[8], and high-performance computing[9] have made real-time virtualization feasible, extending DT from aerospace to diverse industries and, more recently, to healthcare.

The socioeconomic impact of DT adoption is underscored by its market valuation, which surged from USD 10.1 billion in 2023 to an anticipated USD 110.1 billion by 2028, reflecting a compound annual growth rate of 61.3%[10]. Industries ranging from aerospace to healthcare now leverage DT for predictive maintenance, operational optimization, and risk mitigation. Empirical studies have demonstrated that DT implementations can significantly improve operational efficiency in industrial equipment maintenance[2] and enhance surgical outcomes through patient-specific modeling in healthcare[11]. However, these technological advancements are accompanied by persistent challenges in interoperability, data security, and model fidelity that continue to shape research priorities[12-14].

This review traces the historical trajectory of DT from its aerospace origins to its current role in precision medicine. Special emphasis is placed on clinical applications in cardiology, oncology, and chronic disease, where DT supports patient-specific modeling and treatment optimization. Key challenges and future opportunities, including explainable AI, federated learning, and ethical governance, are also discussed.

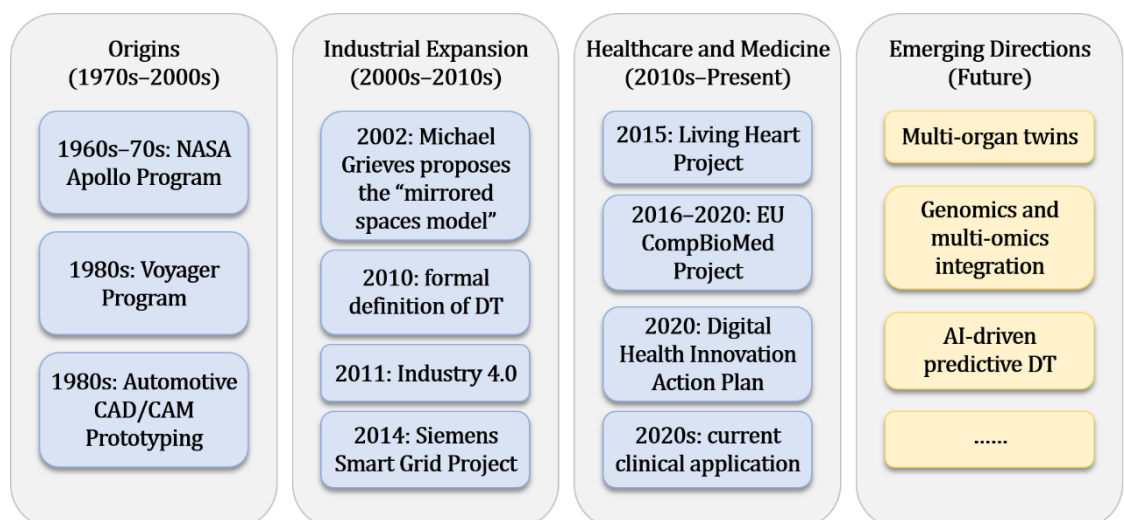

Fig.1 Developmental timeline of digital twin technology.

**Table 1. Representative Clinical Applications of Digital Twin in Medicine**

| Area | Example of Application | Benefits | Limitations |
|---|---|---|---|
| Cardiology | Living Heart Project[32]; heart failure decompensation prediction[62] | Personalized treatment planning; early risk detection | Requires multimodal data and high model fidelity |
| Oncology | Tumor progression modeling[41]; radiotherapy planning[47] | Tracks heterogeneity; optimizes therapy | Data heterogeneity; limited large-scale validation |
| Neurology | Parkinson's DT integrating gait, imaging, and genetics[43] | Forecasts disease trajectory; guides early intervention | Lack of standardized biomarkers; variability |
| Pharmacology | PBPK modeling[44]; virtual microdosing trials[45] | Accelerates drug discovery; predicts toxicity | Regulatory acceptance is still evolving |
| Clinical Trials | Virtual patient cohorts for rare cancers[60] | Improves trial design and recruitment efficiency | Requires robust data integration and validation |
| Chronic Care | Diabetes DT with glucose monitoring[63] | Continuous adaptive monitoring; supports coordinated care | Integration into workflows remains challenging |

**Origins and Early Development (1970s–2000s)**

The conceptual origins of DT technology can be traced to NASA's groundbreaking Apollo program during the 1960s–70s, where engineers pioneered the use of mirrored systems for spacecraft simulation and mission troubleshooting[3]. A seminal moment occurred during the Apollo 13 crisis in April 1970, when Mission Control engineers famously employed complete physical duplicates of the spacecraft's systems to meticulously simulate potential failure scenarios and develop emergency solutions in real time[15]. These full-scale physical mockups, although technologically primitive by today's standards, embodied the fundamental principles that would later define DT technology: the creation of identical counterparts to enable remote diagnostics, predictive analysis, and solution testing in a risk-free environment[9]. Although these early implementations relied on physical rather than digital replication, they established the conceptual framework that would eventually evolve into modern DT systems through subsequent technological advancements in computing, sensor technology, and data analytics. This pioneering work by NASA engineers not only aided the Apollo 13 mission but also laid the theoretical and practical foundation for what would become one of the most transformative technologies of the 21st century[16].

In the 1980s and 1990s, digital simulation advanced in both aerospace and manufacturing. At NASA, hybrid models integrating telemetry data were used to predict spacecraft performance, while in the automotive sector, computer-aided design and manufacturing (CAD/CAM) enabled virtual prototyping that reduced reliance on physical crash tests[17]. These developments highlighted the need to bridge virtual design with real-world performance.

Parallel developments occurred in the manufacturing sector during this period, with different motivations and implementations. During the 1980s, the adoption of computer-integrated manufacturing by the automotive industry laid the groundwork for early DT practices on production lines[12]. General Motors' deployment of CAD/CAM technologies enabled virtual prototyping and simulation, which in turn helped to reduce the frequency and cost of physical crash tests, resulting in notable efficiency gains throughout its vehicle design pipeline[18]. However, these models remained firmly confined to the design phase, lacking any meaningful integration with operational data from fielded products. During the early stages of automotive simulation development, a fundamental challenge persisted: virtual design models often failed to fully capture the complexities of real-world physical performance. This gap limited the accuracy and reliability of simulations, highlighting the ongoing need to better integrate virtual prototyping with actual vehicle behavior.

The theoretical framework for modern DT coalesced at the University of Michigan in the early 2000s through the work of Michael Grieves on product lifecycle management[3]. Grieves' "mirrored spaces model" articulated the three core components that define all contemporary DT: the physical entity in real space, the virtual counterpart in virtual space, and the data linkage that connects them. This bidirectional framework allowed simulations not only to reflect but also to influence physical systems. In 2010, NASA's Modeling and Simulation Roadmap formally defined DT as "integrated multiphysics, multiscale, probabilistic simulations of complex systems," cementing its role as a methodological cornerstone[2].

**Expansion Across Industries (2000s–2010s)**

The first decade of the 21st century marked a turning point as DT expanded beyond aerospace into other industries, driven by the convergence of enabling technologies such as IoT sensors, cloud computing, and machine learning[19]. These advances addressed the limitations of early simulation, including restricted computational power, lack of real-time data, and rigid models[20]. This period saw DT technology evolve into three distinct architectural paradigms tailored to various industrial applications. Component-level twins became prominent in capital-intensive sectors where the value of machinery justified the computational complexity involved. These DT systems typically integrate real-time sensor data monitoring critical parameters such as temperature, pressure, and vibration, and compare operational data against extensive libraries of simulated failure modes[2]. Such frameworks demonstrated strong predictive maintenance capabilities, detecting potential failures well in advance, often dozens of operational hours before occurrence and enabling proactive interventions that significantly reduce unplanned downtime[21].

System-level twins gained traction in distributed infrastructure applications during this period. Siemens' Smart Grid Solutions demonstrated this approach through their implementation in Barcelona's power network, where they developed a digital replica of the electrical distribution system that integrated operational data from multiple sources, including smart meters and equipment sensors[22]. This project is an early example of how DT could coordinate complex utility networks, although quantitative claims require verification from primary sources.

Perhaps the most transformative development during this period was the emergence of process-level twins in manufacturing environments. The Industry 4.0 movement was pioneered in Germany's manufacturing sector, establishing DT as a cornerstone technology for smart production systems[23,24]. This paradigm shift demonstrated how virtual replicas can fundamentally transform traditional manufacturing through bidirectional synchronization between physical operations and computational models[25]. The integration of discrete event simulation with IoT-enabled material flow monitoring enabled the creation of closed-loop control systems[26]. DT platforms can dynamically reconcile planned production schedules with real-time operational data across multiple manufacturing units, facilitating adaptive rescheduling of maintenance activities and workforce deployment[2]. Evidence from industry case studies and analyst reports frequently suggests a trend toward improved overall equipment effectiveness (OEE) when compared to conventional planning, though the extent of improvement is highly context-dependent[27]. This highlights the transformative potential of process-level DT in achieving responsive, self-optimizing production environments.

The economic impact of industrial DT applications became increasingly evident during this period. Industry analyses began to report consistent improvements in key performance metrics such as maintenance efficiency, production throughput, and quality monitoring[12,25]. These demonstrable benefits contributed to rapid market expansion and widespread institutional adoption. By the late 2010s, major industrial players had established dedicated DT initiatives[27], signaling a shift from conceptual exploration to strategic investment based on measurable returns.

**Entry into Healthcare and Medicine (2010s–Present)**

The healthcare sector's adoption of DT technology has followed a distinct evolutionary path compared with industrial applications, constrained by multifaceted regulatory hurdles and the profound complexity inherent in biological systems[28]. Interestingly, the genesis of medical DT did not originate from direct clinical demands, but rather emerged as an unexpected byproduct of aerospace research. NASA's pioneering Human Research Program initially developed sophisticated computational physiology models to predict and mitigate astronaut health risks during extended space missions[29]. These comprehensive models, originally engineered to simulate the intricate effects of microgravity on human physiology, inadvertently laid the crucial groundwork for terrestrial medical applications[30]. A pivotal transition occurred when interdisciplinary research teams recognized the transformative potential of these space-derived models for creating highly accurate, patient-specific simulations that could revolutionize precision medicine[31]. This serendipitous cross-pollination between aerospace innovation and medical science ultimately birthed an entirely new paradigm in healthcare technology.

Medical DT technology achieved its first major regulatory milestone in 2015 with the Food and Drug Administration (FDA) clearance of the groundbreaking Living Heart Project, an innovative collaboration between Dassault Systèmes' simulation experts and premier academic cardiology centers[32]. This pioneering effort marked a significant advancement in cardiac modeling by creating highly personalized virtual heart models that integrate multi-modal clinical data such as magnetic resonance imaging-derived anatomical structures, electrophysiological mapping, and hemodynamic measurements. Clinicians were able to use these sophisticated DT to simulate individual responses to antiarrhythmic therapies, facilitating personalized treatment planning[33]. Subsequent validation efforts demonstrated meaningful alignment between simulation-based predictions and observed patient outcomes, contributing to evolving regulatory recognition of computational physiology models as supportive tools in cardiology and underscoring their potential to transform patient-specific therapeutic decision-making[34].

The adoption of DT technology in the healthcare sector has been propelled by three major technological advances that uniquely address its challenges. First, revolutionary multiscale biological modeling frameworks overcame the complexity of human physiology by enabling seamless integration across hierarchical biological systems[11]. These sophisticated computational approaches married finite element analysis for macroscopic tissue mechanics with agent-based modeling for microscopic cellular processes, creating an unprecedented continuum of physiological simulation[35]. Notably, the European Union's CompBioMed project demonstrated the power of this approach through its advanced liver DT, which simulates drug metabolism across molecular, cellular, and organ levels, offering a more comprehensive understanding of pharmacokinetics[30]. Second, the rapid expansion of medical IoT ecosystems encompassing implantable biosensors, wearable diagnostics, and smart hospital equipment enabled continuous acquisition of rich clinical data, facilitating dynamic calibration and real-time updating of DT models[36]. This capability proved to be particularly impactful in cardiology, where devices such as implantable cardiac monitors feed continuous physiological data into adaptive DT to provide personalized and evolving patient assessments. Third, the evolution of regulatory science was equally important, exemplified by the FDA's 2020 Digital Health Innovation Action Plan, which established clear evaluation frameworks and precertification pathways for software-as-a-medical-device, including AI and machine learning–based technologies[37]. These regulatory advancements ensured that complex, adaptive systems like DT could be validated and monitored post-market while maintaining rigorous patient safety standards, thus enabling their effective clinical deployment.

**Current Clinical Applications**

DT technology is currently revolutionizing healthcare by enabling dynamic, patient-specific models that evolve in real time with disease progression. By continuously integrating multimodal data, including medical imaging, liquid biopsies, and real-time physiological monitoring, DT generates comprehensive, temporally coherent representations of individual health, capable of detecting subtle pathological changes that are often missed by conventional methods[38-40]. In oncology, DT has demonstrated the ability to track tumor heterogeneity and predict metastatic potential with high precision through longitudinal analysis of serial imaging, facilitating timely interventions and accurate staging[41]. In chronic diseases, these models can capture individual variability in progression; for example, in heart failure, the models integrate imaging and biomarker data to predict decompensation risk days in advance[11,42],, whereas in neurodegenerative diseases such as Parkinson's disease, DT models can combine gait analysis, neuroimaging, and genetic factors to model disease trajectories and guide early therapeutic decisions[43]. Overall, DT provides the capability to shift healthcare from reactive treatment to proactive, personalized management.

In therapeutic optimization, DT enables clinicians to simulate thousands of treatment permutations, including drug regimens, surgical plans, and radiation protocols, on the basis of personalized anatomical and physiological data. Physiologically-based pharmacokinetic (PBPK) models incorporate dynamic inputs such as positron emission tomography (PET) microdosing and pathology data to predict drug distribution, efficacy, and toxicity, supporting tailored regimens[44]. For instance, DT has been applied to predict salvage therapies in non-small cell lung cancer after immunotherapy progression[45]. In radiation therapy, patient-specific DT improves dosimetry precision by simulating radiopharmaceutical interactions and optimizing protocols[46,47]. Surgical planning benefits from anatomical DT that allows virtual procedure simulation and prosthesis testing, enhancing accuracy in interventions such as transcatheter aortic valve replacement and orthopedic implant alignment[48-50].

Beyond clinical care, DT has transformed drug discovery by providing *in silico* platforms that complement traditional pipelines, shortening development timelines, and improving safety predictions[51,52]. Additionally, they have facilitated high-throughput absorption, distribution, metabolism, excretion, and toxicity screening and simulated Phase 0 microdosing trials across virtual populations, including special physiological states such as pregnancy[53-55]. In the field of theranostics, DT frameworks combining molecular imaging data with pharmacokinetic simulations have been explored to accelerate the development of novel PET tracers and radionuclide-labeled drugs, thereby bridging early drug discovery with translational nuclear medicine. DT can also improve drug design by modeling hepatic metabolism variability to predict efficacy and toxicity, reducing late-stage failures[56,57]. This approach enables the acceleration of candidate evaluation, promotes clinical translation, and reduces costs[58].

Finally, DT reshapes clinical trial design by generating virtual patient cohorts that simulate treatment effects, optimize recruitment, and enhance trial power. This is particularly valuable in rare diseases and early-phase oncology trials, where recruitment is challenging[59,60]. By integrating multimodal clinical, genomic, radiomic, and socioeconomic data, virtual cohorts reflect real-world patient diversity, enabling refined protocol design to minimize failures and ethical concerns[61].

**Challenges and Limitations**

As DT technology matures from conceptual frameworks to early-stage clinical deployment, it is encountering a range of technical, regulatory, and ethical challenges that must be addressed to support safe, effective, and equitable integration into healthcare systems. One of the foremost technical limitations is the lack of interoperability, as current DT platforms often use proprietary formats and disease-specific designs, hindering collaboration and scalability. Addressing this issue requires comprehensive standardization efforts[64].

Another pressing issue is data privacy and security. Healthcare data comprises some of the most sensitive and personal information available, covering medical histories, genetic profiles, and treatment records. Creating DT systems involves collecting and integrating all of this information, and there are serious concerns about data being accessed inappropriately or breaches leading to privacy violations and identity theft[65]. Ensuring secure, anonymized, and consent-driven data pipelines is essential, particularly when data are shared across institutions or used for secondary research purposes[66].

**Future Trajectory**

The continued evolution of DT will rely on overcoming key technical and regulatory challenges. As DT methods increasingly inform clinical decisions, explainable AI frameworks that clearly articulate their reasoning will be essential to maintain clinician trust and regulatory acceptance[67]. Federated learning has emerged as a promising way to enable cross-institutional DT training without centralized data sharing[68]. Additionally, harmonized regulatory frameworks will accelerate the clinical adoption of DT technologies, ensuring they meet standardized safety and efficacy requirements[69]. Looking ahead, the integration of AI and machine learning will further enhance DT capabilities, enabling highly personalized treatment optimization and predictive analytics. By analyzing real-time patient data, genetic profiles, and historical records, AI-powered DT could revolutionize proactive care, predicting disease onset and tailoring interventions to individual needs[70].

**Summary**

DT technology has evolved from NASA's aerospace applications into a transformative tool across industries, including healthcare. In medicine, DT integrates real-time imaging, physiological data, and molecular data to create dynamic, patient-specific models that support early diagnosis, risk prediction, and personalized treatment planning. Current clinical applications span oncology, cardiology, and neurology, where DT is used to simulate therapeutic responses, optimize radiotherapy and pharmacotherapy, and guide complex surgical procedures. In drug development, DT is used to accelerate candidate screening and reduce reliance on traditional clinical trials by enabling high-fidelity *in silico* testing. Despite this progress, key challenges remain, including a lack of interoperability, data privacy concerns, and the need for regulatory clarity. Emerging solutions such as explainable AI, federated learning, and harmonized frameworks offer promising paths forward. In the future, DT is poised to reshape precision medicine by enabling predictive, preventive, and personalized care at scale. Their continued success will depend not only on technical innovation but also on ethical design and multidisciplinary collaboration.

**Clinics Care Points**

DT models create dynamic, patient-specific representations that integrate imaging, physiological, and molecular data for precision diagnosis and treatment.

Clinical applications of DT span cardiology, oncology, and neurology, enabling personalized simulation of disease progression and therapy outcomes.

Integration of DT frameworks into clinical workflows supports proactive monitoring, treatment optimization, and adaptive care.

Successful clinical translation of DT technologies requires multidisciplinary collaboration, robust validation, data interoperability, and sound ethical governance.

**Disclosure**

The authors have no conflicts of interest to declare.

**Funding and additional information**

This work was supported by the Science and Technology Commission of Shanghai Municipality (STCSM) (No. 25TS1405600).